\documentclass[runningheads]{llncs}
\usepackage[numbers]{natbib}

\usepackage[T1]{fontenc}
\usepackage{multirow}
\usepackage[utf8]{inputenc}
\usepackage{graphicx}
\usepackage{amsmath}
\usepackage{array}

\title{A Spanish dataset for Targeted Sentiment Analysis of political headlines}
\titlerunning{Targeted Sentiment Analysis of news headlines}

\author{
    Tomás Alves Salgueiro \inst{1} $^a$ \and
    Emilio Recart Zapata \inst{1, 3} $^a$  \and
    Damián Furman \inst{2} $^b$\and
    Juan Manuel Pérez\inst{2} $^b$ \and
    Pablo Nicolás Fernández Larrosa \inst{1} $^b$
}

\institute{
    Instituto de Fisiología, Biología Molecular y Neurociencia, CONICET, UBA, \\
    \email{\{talvessalgueiro, fernandezlarrosa\}@fbmc.fcen.uba.ar} \\
    \and
    Instituto de Ciencias de la Computación, CONICET, UBA \\
    \email{\{jmperez, dfurman\}@dc.uba.ar}\and
    Facultad de Psicología, UBA
    \\\email{recartemilio@psi.uba.ar}
}
\date{June 2022}

\begin{document}

\maketitle

\def\thefootnote{$^a$, $^b$}\footnotetext{These authors contributed equally to this work}\def\thefootnote{\arabic{footnote}}

\begin{abstract}
    Subjective texts have been especially studied by several works as they can induce certain behaviours in their users. Most work focuses on user-generated texts in social networks, but some other texts also comprise opinions on certain topics and could influence judgement criteria during political decisions. In this work, we address the task of Targeted Sentiment Analysis for the domain of news headlines, published by the main outlets during the 2019 Argentinean Presidential Elections. For this purpose, we present a polarity dataset of 1,976 headlines mentioning candidates in the 2019 elections at the target level. Preliminary experiments with state-of-the-art classification algorithms based on pre-trained linguistic models suggest that target information is helpful for this task. We make our data and pre-trained models publicly available.
\end{abstract}

\section{Introduction}

Extracting opinions from subjective texts has attracted a lot of the Interest since the eclosion of Internet and Social Networks, given the unprecedented availability of opinion-rich resources \cite{pang2008opinion}. Most works for opinion mining  are directed towards user-generated texts from social networks; however, some other texts --such as news headlines-- also convey subjective content about certain topics or entities.

Particularly, it is of interest to analyze the role of the media and campaigns in the formation of judgment criteria during political decisions \cite{kushin2013did}. The rise of social networks marked a great dynamism in the management and transfer of large volumes of data, playing a fundamental role in the transmission of information in political and candidate campaigns at a massive level \cite{reiter2021good}.

The current work is part of a research project to evaluate the cognitive processes underlying the presidential elections, and assessing the impact of greater exposure to positive content in news headlines associated with the candidates. For this purpose, we are interested in analyzing headlines mentioning candidates by the main national written media.

In this paper, we present an approach to the task of Targeted Sentiment Analysis for the domain of newspaper headlines. To the best of our knowledge, no Spanish dataset is available for this task. To bridge this gap, we present a novel dataset of headlines mentioning candidates in the 2019 elections in Argentina, having annotations at target level instead of assigning a single polarity to the whole sentence. Preliminary experiments with state-of-the-art techniques suggest that classifiers that consume both the headline and the target improve their performance for the task over those that only consume the headline, giving indications that both sources of information are useful. We make our dataset and the pre-trained models available for further research.

\section{Previous work}

Sentiment analysis and opinion mining have been one of the most popular applications of NLP, and several workshops were dedicated to this topic \cite{rosenthal-2017-semeval,garcia2020overview}. This task usually consists of the prediction of the polarity of a text (positive, negative, or neutral) or a Likert-scale rating from negative to positive \cite{socher-etal-2013-recursive}. One variant of this problem is \emph{Aspect-based sentiment analysis} (ABSA), in which the prediction  is performed on a text and a particular aspect \cite{pavlopoulos2014aspect}, something useful to extract multiple opinions from a text. Similarly, \emph{Targeted sentiment analysis} analyses the polarity for a given entity mentioned in the text \cite{mitchell2013open}. As far as we know, in spite of the many resources available for Spanish \cite{miranda2017review,garcia2020overview}, no dataset is available for this language in this particular subtask.

Pretrained language models based on transformers \cite{vaswani2017attention} are state-of-the-art for most NLP tasks. BERT \cite{devlin2018bert} and variants \cite{liu2019roberta} are among the most popular classification techniques nowadays and have top performances on language-understanding benchmarks such as GLUE \cite{wang-etal-2018-glue}. Several models have been pre-trained in Spanish \cite{canete2020spanish, delarosa2022bertin,perez2022robertuito}. \citet{nozza2020mask} provide a good overview of pre-trained models across several domains and languages.

\section{Data}

\begin{figure}[t!]
    \centering
    \caption{Examples of the dataset}
    \label{fig:examples}
    \includegraphics[width=\textwidth]{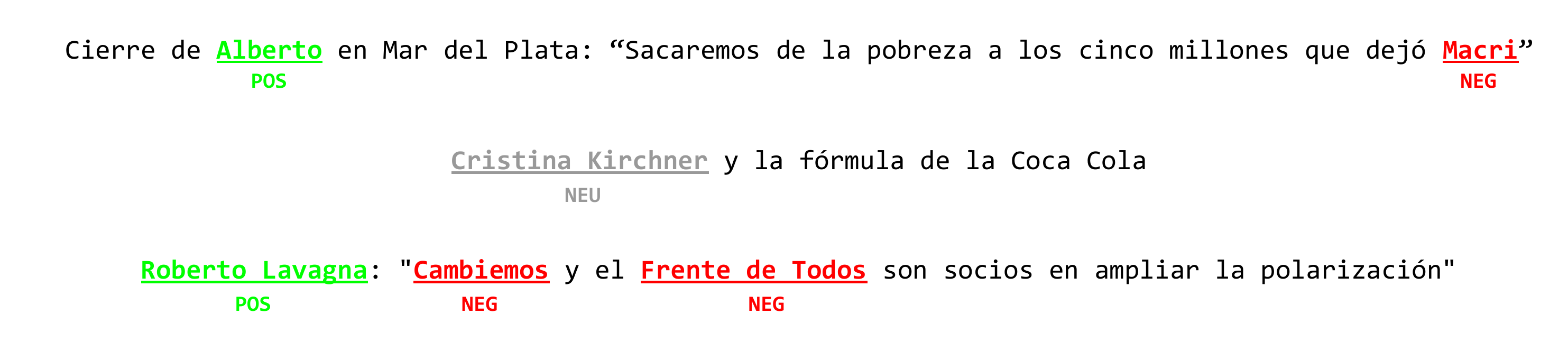}
\end{figure}

We collected news headlines published between 21 September and 27 October 2019 by the main national news outlets: Ambito, Clarin, El Cronista, INFOBAE, La Nacion, Pagina 12, Perfil, Popular, La Izquierda Diario, Prensa Obrera, Tiempo Argentino. We selected only those headlines mentioning one of the contending parties or candidates in the national election.

Three annotators were hired to label the headlines. For each pair of headline and  target (a political party or candidate), the annotator assigned a polarity to the pair. To diminish political biases as much as possible, we masked the targets to the annotators (e.g. Mauricio Macri was shown as \verb|[MASK]|).

Each instance was annotated by the three workers as the task is subjective. The agreement was measured using Krippendorff's Alpha \cite{krippendorff2018content} and turned out to be $\alpha = 0.62$ (moderate/substantial agreement). A majority voting scheme was used to aggregate the labels, discarding those targets for which the three raters assigned different polarities.


The resulting dataset consists of 1,976 headlines and 2,439 targets. 1,567 headlines have exactly one target, and the remaining have two or more. Among these, 165 headlines feature mixed polarities, mostly in the negative/positive form. Figure \ref{fig:examples} illustrates some examples of the dataset.

\section{Method}

\begin{figure}[t]
    \centering
    \caption{Classification models for the task. Target-unaware classifier consumes only the headline, while the target-aware version consumes both headline and target}
    \includegraphics[width=0.5\textwidth]{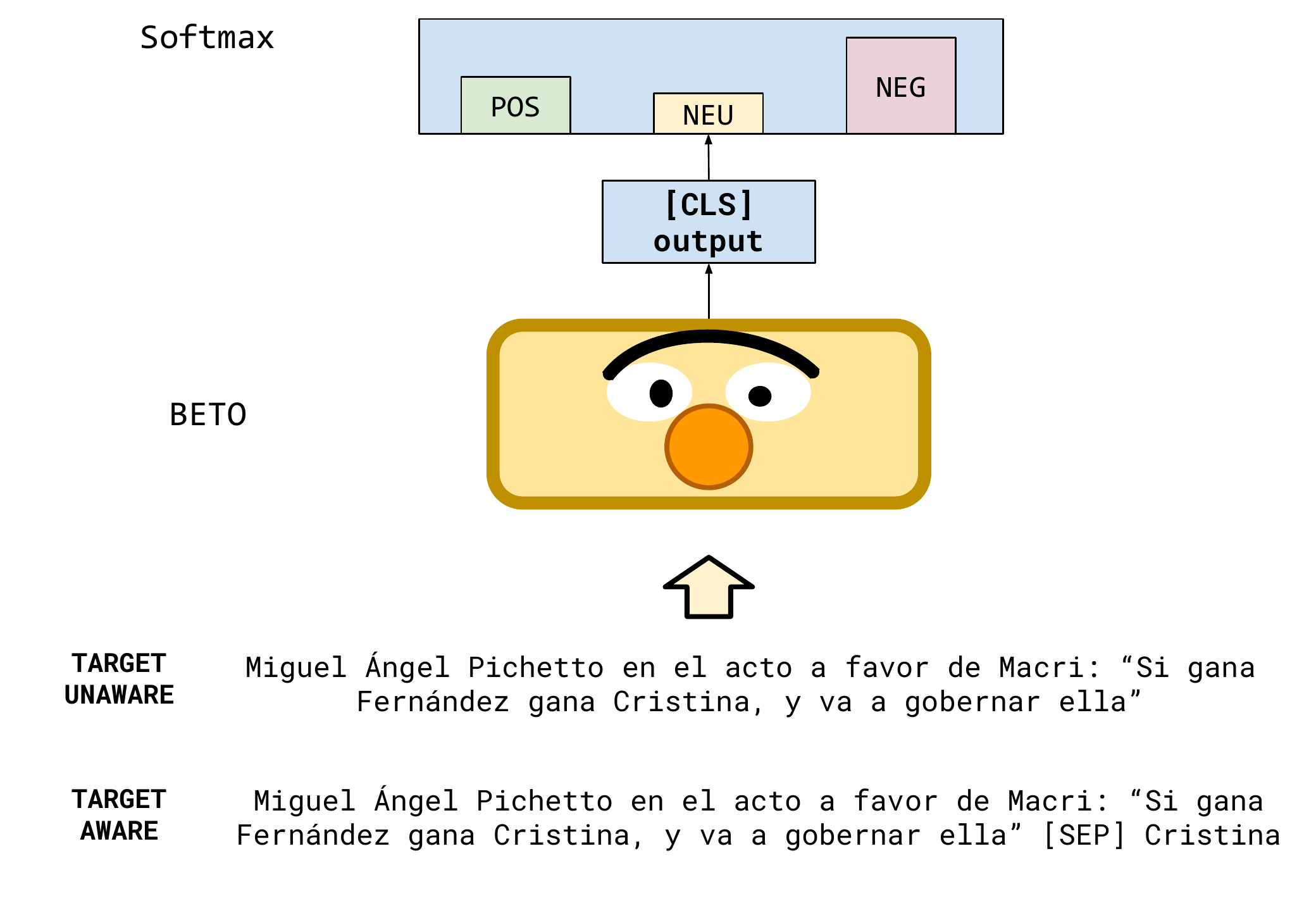}
    \label{fig:classifiers}
\end{figure}

To assess whether algorithms can leverage target information, we proposed a classification experiment with target-aware and target-unaware models:

\begin{enumerate}
    \item Target-unaware: a classifier only consuming the headline and predicting an overall polarity
    \item Target-aware: a classifier consuming both the headline and the target and predicting a polarity for the pair
\end{enumerate}

Our algorithms are based on BETO \cite{canete2020spanish}. The only difference between the two versions is the input: in the target-unaware classifier the input consists of the headline alone, while in the target-aware version it is made of the headline and the target with the special token \verb|[SEP]| in-between them \cite{sun2019utilizing}.

Figure \ref{fig:classifiers} illustrates the two algorithms and sample input. We followed standard BERT training for classification tasks, fine-tuning with LR = $10^{-5}$ for 5 epochs, selecting the best model in terms of Macro F1 for the validation split.

Instead of performing a single train/dev/split, we used a Monte Carlo cross-validation \cite{gorman-bedrick-2019-need} with 15 folds. The splits were performed at the headline level to avoid overestimating the performance.

\section{Results}


\newcolumntype{M}[1]{>{\centering\arraybackslash}m{#1}}

\begin{table}[t]
    \centering
    \footnotesize
    \caption{Results of the classification experiments, expressed as the mean $\pm$ standard deviation of the 15 runs of the experiments.}
    \label{tab:results}

    \begin{tabular}{p{0.12\textwidth} p{0.15\textwidth} M{0.2\textwidth} M{0.2\textwidth}}
                                  &           & \multicolumn{2}{c}{Target}                   \\
        Metric                    &           & Unaware                    & Aware           \\
        \hline
        \multirow{3}{*}{Negative} & Precision & $62.17 \pm 3.7$            & $70.23 \pm 3.2$ \\
                                  & Recall    & $62.04 \pm 4.1$            & $69.77 \pm 3.7$ \\
                                  & F1        & $62.01 \pm 3.0$            & $69.93 \pm 2.6$ \\
        \hline
        \multirow{3}{*}{Neutral}  & Precision & $59.62 \pm 3.9$            & $61.04 \pm 4.8$ \\
                                  & Recall    & $55.59 \pm 8.6$            & $57.09 \pm 7.3$ \\
                                  & F1        & $57.07 \pm 5.0$            & $58.54 \pm 3.5$ \\
        \hline
        \multirow{3}{*}{Positive} & Precision & $65.95 \pm 3.9$            & $70.36 \pm 3.9$ \\
                                  & Recall    & $68.97 \pm 4.3$            & $73.62 \pm 5.4$ \\
                                  & F1        & $67.25 \pm 2.0$            & $71.73 \pm 2.4$ \\
        \hline
        \multirow{3}{*}{Macro}    & Precision & $62.58 \pm 1.7$            & $67.21 \pm 1.3$ \\
                                  & Recall    & $62.20 \pm 2.0$            & $66.82 \pm 1.4$ \\
                                  & F1        & $62.11 \pm 2.1$            & $66.73 \pm 1.4$ \\
        \hline
    \end{tabular}
\end{table}

Table \ref{tab:results} displays the results of the classification experiments, expressed as the mean and its standard deviation for the 15 runs. For all metrics, the target aware version performs above the target unaware model.

These differences are statistically significant ($p < 0.001$, one-sided Wilcoxon Mann-Whitney U-test, FDR corrected) \cite{wilcoxon1992individual} for all cases but for the neutral class. This is in line with our data: as neutral targets mostly do not mix with positive or negative polarities, predicting a general polarity is good enough.

\section{Conclusion and future work}

We presented a dataset for targeted sentiment analysis in Spanish news headlines, showing that state-of-the-art classifiers can leverage its target information. We make the dataset available at the huggingface hub \footnote{\url{pysentimiento/spanish-targeted-sentiment-headlines}}.

As future work, we plan to enlarge the dataset with new sources and other events, and explore new classification techniques.
%
%
%
\bibliographystyle{IEEEtranN}
\bibliography{biblio}

\begin{thebibliography}{21}
\providecommand{\natexlab}[1]{#1}
\providecommand{\url}[1]{#1}
\csname url@samestyle\endcsname
\providecommand{\newblock}{\relax}
\providecommand{\bibinfo}[2]{#2}
\providecommand{\BIBentrySTDinterwordspacing}{\spaceskip=0pt\relax}
\providecommand{\BIBentryALTinterwordstretchfactor}{4}
\providecommand{\BIBentryALTinterwordspacing}{\spaceskip=\fontdimen2\font plus
\BIBentryALTinterwordstretchfactor\fontdimen3\font minus
  \fontdimen4\font\relax}
\providecommand{\BIBforeignlanguage}[2]{{%
\expandafter\ifx\csname l@#1\endcsname\relax
\typeout{** WARNING: IEEEtranN.bst: No hyphenation pattern has been}%
\typeout{** loaded for the language `#1'. Using the pattern for}%
\typeout{** the default language instead.}%
\else
\language=\csname l@#1\endcsname
\fi
#2}}
\providecommand{\BIBdecl}{\relax}
\BIBdecl

\bibitem[Pang and Lee(2008)]{pang2008opinion}
\BIBentryALTinterwordspacing
B.~Pang and L.~Lee, ``Opinion mining and sentiment analysis,'' \emph{Found.
  Trends Inf. Retr.}, vol.~2, no. 1–2, p. 1–135, jan 2008. [Online].
  Available: \url{https://doi.org/10.1561/1500000011}
\BIBentrySTDinterwordspacing

\bibitem[Kushin and Yamamoto(2013)]{kushin2013did}
M.~J. Kushin and M.~Yamamoto, ``Did social media really matter? college
  students’ use of online media and political decision making in the 2008
  election,'' in \emph{New Media, Campaigning and the 2008 Facebook
  Election}.\hskip 1em plus 0.5em minus 0.4em\relax Routledge, 2013, pp.
  63--86.

\bibitem[Reiter and Matthes(2021)]{reiter2021good}
F.~Reiter and J.~Matthes, ``“the good, the bad, and the ugly”: A panel
  study on the reciprocal effects of negative, dirty, and positive campaigning
  on political distrust,'' \emph{Mass Communication and Society}, pp. 1--24,
  2021.

\bibitem[Rosenthal et~al.(2017)Rosenthal, Farra, and
  Nakov]{rosenthal-2017-semeval}
\BIBentryALTinterwordspacing
S.~Rosenthal, N.~Farra, and P.~Nakov, ``{S}em{E}val-2017 task 4: Sentiment
  analysis in {T}witter,'' in \emph{Proceedings of the 11th International
  Workshop on Semantic Evaluation ({S}em{E}val-2017)}.\hskip 1em plus 0.5em
  minus 0.4em\relax Vancouver, Canada: Association for Computational
  Linguistics, Aug. 2017, pp. 502--518. [Online]. Available:
  \url{https://aclanthology.org/S17-2088}
\BIBentrySTDinterwordspacing

\bibitem[Garc{\'\i}a-Vegaa et~al.(2020)Garc{\'\i}a-Vegaa, D{\'\i}az-Galianoa,
  Garc{\'\i}a-Cumbrerasa, del Arcoa, Montejo-R{\'a}eza, Jim{\'e}nez-Zafraa,
  C{\'a}marab, Aguilarc, Antonio, Cabezudod, et~al.]{garcia2020overview}
M.~Garc{\'\i}a-Vegaa, M.~C. D{\'\i}az-Galianoa, M.~{\'A}.
  Garc{\'\i}a-Cumbrerasa, F.~M.~P. del Arcoa, A.~Montejo-R{\'a}eza, S.~M.
  Jim{\'e}nez-Zafraa, E.~M. C{\'a}marab, C.~A. Aguilarc, M.~Antonio,
  S.~Cabezudod \emph{et~al.}, ``Overview of tass 2020: introducing emotion
  detection,'' 2020.

\bibitem[Socher et~al.(2013)Socher, Perelygin, Wu, Chuang, Manning, Ng, and
  Potts]{socher-etal-2013-recursive}
\BIBentryALTinterwordspacing
R.~Socher, A.~Perelygin, J.~Wu, J.~Chuang, C.~D. Manning, A.~Ng, and C.~Potts,
  ``Recursive deep models for semantic compositionality over a sentiment
  treebank,'' in \emph{Proceedings of the 2013 Conference on Empirical Methods
  in Natural Language Processing}.\hskip 1em plus 0.5em minus 0.4em\relax
  Seattle, Washington, USA: Association for Computational Linguistics, Oct.
  2013, pp. 1631--1642. [Online]. Available:
  \url{https://aclanthology.org/D13-1170}
\BIBentrySTDinterwordspacing

\bibitem[Pavlopoulos(2014)]{pavlopoulos2014aspect}
I.~Pavlopoulos, ``Aspect based sentiment analysis,'' \emph{Athens University of
  Economics and Business}, 2014.

\bibitem[Mitchell et~al.(2013)Mitchell, Aguilar, Wilson, and
  Van~Durme]{mitchell2013open}
M.~Mitchell, J.~Aguilar, T.~Wilson, and B.~Van~Durme, ``Open domain targeted
  sentiment,'' in \emph{Proceedings of the 2013 Conference on Empirical Methods
  in Natural Language Processing}, 2013, pp. 1643--1654.

\bibitem[Miranda and Guzman(2017)]{miranda2017review}
C.~H. Miranda and J.~Guzman, ``A review of sentiment analysis in spanish,''
  \emph{Tecciencia}, vol.~12, no.~22, pp. 35--48, 2017.

\bibitem[Vaswani et~al.(2017)Vaswani, Shazeer, Parmar, Uszkoreit, Jones, Gomez,
  Kaiser, and Polosukhin]{vaswani2017attention}
A.~Vaswani, N.~Shazeer, N.~Parmar, J.~Uszkoreit, L.~Jones, A.~N. Gomez,
  L.~Kaiser, and I.~Polosukhin, ``Attention is all you need,'' \emph{arXiv
  preprint arXiv:1706.03762}, 2017.

\bibitem[Devlin et~al.(2019)Devlin, Chang, Lee, and Toutanova]{devlin2018bert}
\BIBentryALTinterwordspacing
J.~Devlin, M.-W. Chang, K.~Lee, and K.~Toutanova, ``{BERT}: Pre-training of
  deep bidirectional transformers for language understanding,'' pp. 4171--4186,
  Jun. 2019. [Online]. Available: \url{https://aclanthology.org/N19-1423}
\BIBentrySTDinterwordspacing

\bibitem[Liu et~al.(2019)Liu, Ott, Goyal, Du, Joshi, Chen, Levy, Lewis,
  Zettlemoyer, and Stoyanov]{liu2019roberta}
Y.~Liu, M.~Ott, N.~Goyal, J.~Du, M.~Joshi, D.~Chen, O.~Levy, M.~Lewis,
  L.~Zettlemoyer, and V.~Stoyanov, ``Roberta: A robustly optimized bert
  pretraining approach,'' \emph{arXiv preprint arXiv:1907.11692}, 2019.

\bibitem[Wang et~al.(2018)Wang, Singh, Michael, Hill, Levy, and
  Bowman]{wang-etal-2018-glue}
\BIBentryALTinterwordspacing
A.~Wang, A.~Singh, J.~Michael, F.~Hill, O.~Levy, and S.~Bowman, ``{GLUE}: A
  multi-task benchmark and analysis platform for natural language
  understanding,'' in \emph{Proceedings of the 2018 {EMNLP} Workshop
  {B}lackbox{NLP}: Analyzing and Interpreting Neural Networks for {NLP}}.\hskip
  1em plus 0.5em minus 0.4em\relax Brussels, Belgium: Association for
  Computational Linguistics, Nov. 2018, pp. 353--355. [Online]. Available:
  \url{https://aclanthology.org/W18-5446}
\BIBentrySTDinterwordspacing

\bibitem[Canete et~al.(2020)Canete, Chaperon, Fuentes, and
  P{\'e}rez]{canete2020spanish}
J.~Canete, G.~Chaperon, R.~Fuentes, and J.~P{\'e}rez, ``Spanish pre-trained
  bert model and evaluation data,'' \emph{PML4DC at ICLR}, vol. 2020, 2020.

\bibitem[Rosa et~al.(2022)Rosa, Ponferrada, Romero, Villegas, de~Prado~Salas,
  and Grandury]{delarosa2022bertin}
\BIBentryALTinterwordspacing
J.~D.~L. Rosa, E.~G. Ponferrada, M.~Romero, P.~Villegas, P.~G. de~Prado~Salas,
  and M.~Grandury, ``Bertin: Efficient pre-training of a spanish language model
  using perplexity sampling,'' \emph{Procesamiento del Lenguaje Natural},
  vol.~68, no.~0, pp. 13--23, 2022. [Online]. Available:
  \url{http://journal.sepln.org/sepln/ojs/ojs/index.php/pln/article/view/6403}
\BIBentrySTDinterwordspacing

\bibitem[Pérez et~al.(2022)Pérez, Furman, Alonso~Alemany, and
  Luque]{perez2022robertuito}
\BIBentryALTinterwordspacing
J.~M. Pérez, D.~A. Furman, L.~Alonso~Alemany, and F.~M. Luque, ``Robertuito: a
  pre-trained language model for social media text in spanish,'' in
  \emph{Proceedings of the Language Resources and Evaluation Conference}.\hskip
  1em plus 0.5em minus 0.4em\relax Marseille, France: European Language
  Resources Association, June 2022, pp. 7235--7243. [Online]. Available:
  \url{https://aclanthology.org/2022.lrec-1.785}
\BIBentrySTDinterwordspacing

\bibitem[Nozza et~al.(2020)Nozza, Bianchi, and Hovy]{nozza2020mask}
D.~Nozza, F.~Bianchi, and D.~Hovy, ``What the [mask]? making sense of
  language-specific bert models,'' \emph{arXiv preprint arXiv:2003.02912},
  2020.

\bibitem[Krippendorff(2018)]{krippendorff2018content}
K.~Krippendorff, \emph{Content analysis: An introduction to its
  methodology}.\hskip 1em plus 0.5em minus 0.4em\relax Sage publications, 2018.

\bibitem[Sun et~al.(2019)Sun, Huang, and Qiu]{sun2019utilizing}
C.~Sun, L.~Huang, and X.~Qiu, ``Utilizing bert for aspect-based sentiment
  analysis via constructing auxiliary sentence,'' in \emph{Proceedings of
  NAACL-HLT}, 2019, pp. 380--385.

\bibitem[Gorman and Bedrick(2019)]{gorman-bedrick-2019-need}
\BIBentryALTinterwordspacing
K.~Gorman and S.~Bedrick, ``We need to talk about standard splits,'' in
  \emph{Proceedings of the 57th Annual Meeting of the Association for
  Computational Linguistics}.\hskip 1em plus 0.5em minus 0.4em\relax Florence,
  Italy: Association for Computational Linguistics, Jul. 2019, pp. 2786--2791.
  [Online]. Available: \url{https://aclanthology.org/P19-1267}
\BIBentrySTDinterwordspacing

\bibitem[Wilcoxon(1992)]{wilcoxon1992individual}
F.~Wilcoxon, ``Individual comparisons by ranking methods,'' in
  \emph{Breakthroughs in statistics}.\hskip 1em plus 0.5em minus 0.4em\relax
  Springer, 1992, pp. 196--202.

\end{thebibliography}

\end{document}